\documentclass[letterpaper, 10 pt, conference]{ieeeconf}  
\usepackage{url}
\usepackage{graphicx}

\usepackage{graphicx}
\usepackage{float}
\usepackage{array}
\usepackage{caption}
\usepackage{subcaption}
\usepackage{afterpage}
\usepackage{placeins}
\usepackage{amsmath}
\usepackage{algorithm}
\usepackage{algpseudocode}
\usepackage{amsfonts}
\usepackage{amsmath}
\usepackage{cite}
\usepackage[breaklinks]{hyperref}
\usepackage{array}
\usepackage{tabularx}
\usepackage[table]{xcolor}
\usepackage{graphicx}
\usepackage{subcaption}
\usepackage{caption}
\usepackage{float} 
\usepackage{multirow}

\DeclareMathOperator*{\argmin}{arg\,min}

\usepackage[left=0.75in,right=0.75in,top=1in,bottom=0.75in]{geometry}

\IEEEoverridecommandlockouts                              

\title{\LARGE \bf
Learning Soft Driving Constraints from Vectorized Scene \\ Embeddings while Imitating Expert Trajectories  
}

\author{Niloufar Saeidi Mobarakeh$^{1}$, Behzad Khamidehi$^{2}$, Chunlin Li$^{3}$,  Hamidreza Mirkhani$^{2}$, Fazel Arasteh$^{2}$, \\ Mohammed Elmahgiubi$^{2}$, Weize Zhang$^{2}$, Kasra Rezaee$^{2}$, and Pascal Poupart$^{4}$
\thanks{$^{1}$ Simon Fraser University. Work done while at Noah’s Ark Lab., Huawei Technologies, Canada. {\tt\small nsa171@sfu.ca}}
\thanks{$^{2}$Noah’s Ark Lab., Huawei Technologies, Markham, Ontario, Canada. {\tt\small {firstname.lastname}@huawei.com}}
\thanks{$^{3}$ University of Toronto. Work done while at Noah’s Ark Lab., Huawei Technologies, Canada. {\tt\small edwardphi.li@mail.utoronto.ca}}
\thanks{$^{4}$University of Waterloo and Vector Institute, Ontario, Canada. {\tt\small ppoupart@uwaterloo.ca}%
}}

\begin{document}

\setlength{\tabcolsep}{2pt}

\maketitle
\thispagestyle{empty}
\pagestyle{empty}

\begin{abstract}
The primary goal of motion planning is to generate safe and efficient trajectories for vehicles. Traditionally, motion planning models are trained using imitation learning to mimic the behavior of human experts. However, these models often lack interpretability and fail to provide clear justifications for their decisions. We propose a method that integrates constraint learning into imitation learning by extracting driving constraints from expert trajectories. Our approach utilizes vectorized scene embeddings that capture critical spatial and temporal features, enabling the model to identify and generalize constraints across various driving scenarios. We formulate the constraint learning problem using a maximum entropy model, which scores the motion planner's trajectories based on the similarity to the expert trajectory. By separating the scoring process into distinct reward and constraint streams, we improve both the interpretability of the planner’s behavior and its attention to relevant scene components. Unlike existing constraint learning methods that rely on simulators and are typically embedded in reinforcement learning (RL) or inverse reinforcement learning (IRL) frameworks, our method operates without simulators, making it applicable to a wider range of datasets and real-world scenarios. Experimental results on the InD and TrafficJams datasets demonstrate that incorporating driving constraints not only enhances model interpretability but also improves closed-loop performance.

\end{abstract}


\section{INTRODUCTION}

Motion planning involves generating safe and feasible trajectories for vehicles to navigate complex traffic scenarios. These trajectories must strictly adhere to traffic regulations and avoid safety-critical incidents, such as collisions with other vehicles, pedestrians, or maintaining unsafe proximity to road boundaries \cite{paden2016survey,chai2019multipath, chen2023end}. A key challenge in this domain is ensuring that the planned trajectories are safe, especially in scenarios with unpredictable traffic participants. Imitation learning (IL) techniques have gained prominence in addressing this challenge by focusing on replicating human driving behaviors based on real-world data \cite{chen2019deep, le2022survey, lu2023imitation}. However, IL approaches often lack transparency, making it difficult to interpret how and why the model makes certain decisions. This lack of interpretability can be problematic when deploying autonomous driving systems in safety-critical settings, where understanding model behavior is crucial for trust and validation. One promising approach to overcome this limitation is to derive driving constraints directly from observed human behaviors. These constraints—such as maintaining appropriate distances, obeying speed limits, and adhering to right-of-way rules—are intuitive for human drivers but are challenging to encode explicitly in autonomous systems. Human drivers naturally internalize these constraints through experience and context-awareness, whereas autonomous systems require more formalized representations. Capturing these implicit driving rules and encoding them into autonomous systems could enhance both the safety and interpretability of motion planning models, improving their closed-loop performance in real-world scenarios.

\begin{figure}[t]
    \centering
    \includegraphics[trim={0cm 0 0cm 0}, clip,  width=\linewidth]{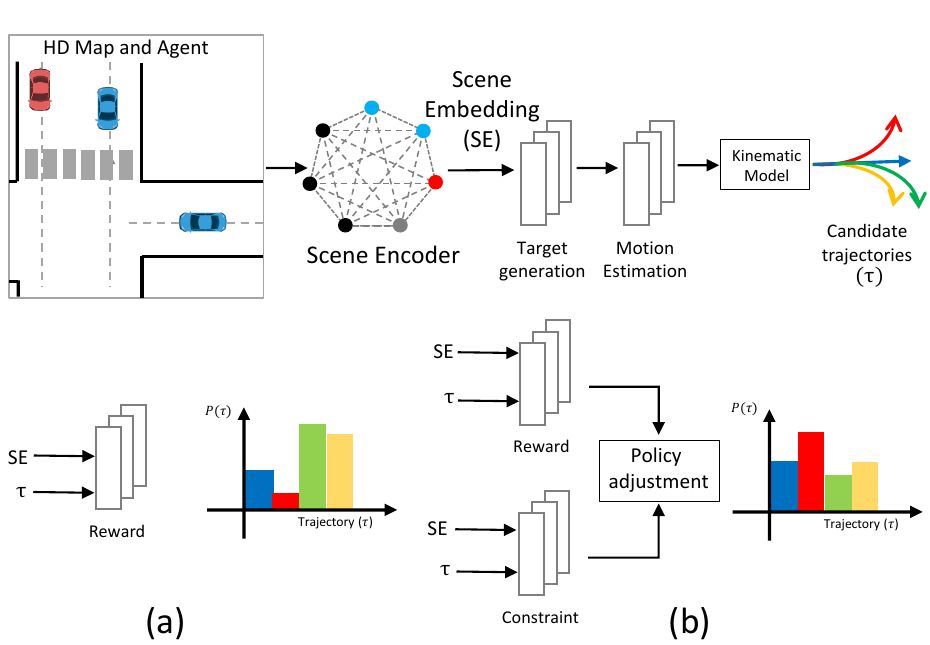}
    \caption{The architecture of our planner. (a) a reward only scheme, where the trajectories are scored based on the similarity to the expert trajectory. (b) Proposed combined reward and constraint learning scheme, where the model learns the constraint values along with the rewards.}
    \label{fig:architecture}
\end{figure}


Several studies have been conducted to explore the constraint learning problem \cite{liu2022benchmarking, howtonotdrive, chou2020learning, scobee2019maximum, icml, pascal_icl, kim2023learning, wang2024large, hu2024long, wang2024safe}. In \cite{scobee2019maximum}, Scobee \textit{et al.} reformulate an Inverse Reinforcement Learning (IRL) on Markov decision processes (MDPs) to estimate state, action, and feature constraints motivating agent behavior. Utilizing the maximum entropy IRL framework of \cite{ziebart2008maximum}, they proposed an iterative method to infer the maximum likelihood constraints. These constraints are then used to explain the model behavior. Malik \textit{et al.} \cite{icml} extend this by focusing on learning hard constraints that a constraint-abiding agent follows. They introduced an algorithm to acquire such constraints. A key distinction between their work and \cite{scobee2019maximum} is that Malik's approach can learn arbitrary Markovian constraints in high-dimensional environments, a capability not supported by the former.
In \cite{pascal_icl}, Gaurav \textit{et al.} propose an Inverse Constraint Learning (ICL) framework to learn soft constraints from demonstrations, assuming that the reward function is given. This approach is similar to IRL, except it focuses on learning the constraints rather than the reward. Similarly, Kim \textit{et al.} \cite{kim2023learning} extend the ICL to multi-task settings to infer shared safety constraints across multiple tasks from demonstrations. The authors in \cite{liu2022benchmarking} develop a benchmark for Inverse Constrained Reinforcement Learning (ICRL) algorithms through synthetic and real-world inspired tasks. These tasks are designed to challenge the ability of ICRL algorithms to accurately infer environmental constraints from observed expert behaviors. The proposed algorithms are tested in a variety of environments to feature a variety of constraint types, including navigation and motion constraints that mimic real-world limitations.

For safety-critic applications like autonomous driving, it is crucial to imitate human drivers and learn the implicit rules that human drivers follow \cite{vitelli2022safetynet}.  
Several studies tried to learn a constraint from human demonstrations and integrate it into autonomous driving frameworks \cite{liu2022benchmarking, lindner2023learning, howtonotdrive, basich2023learning, hu2024long}. For instance, Liu \textit{et al.} \cite{liu2022benchmarking} propose an inverse constraint reinforcement learning benchmark that evaluates constraint inference on a highway driving scenario. 
Lindner \textit{et al.} \cite{lindner2023learning} propose convex constraint learning for reinforcement learning (CoCoRL) and evaluate their method on a 2D driving simulator \cite{highwayenv}. In contrast to \cite{pascal_icl} that assumes a given reward, they learn constraints from demonstrations with unknown rewards. 
Rezaee \textit{et al.} \cite{howtonotdrive} extend the work of \cite{scobee2019maximum} by training a variational auto-encoder (VAE) based model that predicts driving constraints. In \cite{hu2024long}, the authors introduced a dual-constraint mechanism to handle long-term and short-term safety concerns to ensure robust safety during training and deployment.



Existing approaches predominantly rely on interactions with simulators or the availability of reward signals, as they are typically formulated within reinforcement learning (RL) or inverse reinforcement learning (IRL) frameworks. In these paradigms, agents optimize their actions by interacting with the environment. However, developing accurate and robust simulators is particularly challenging, especially in safety-critical domains such as autonomous driving. This complexity arises from the need to model highly dynamic, multi-agent environments with diverse traffic scenarios, rare safety-critical events, and unpredictable behaviors of human drivers. Simulators must also accurately capture the intricate physics of vehicle dynamics, weather conditions, and sensor noise, which can be difficult to replicate in a virtual setting. In contrast, we integrate constraint learning directly within the imitation learning framework, thereby obviating the need for intricate simulator design or access to reward signals. Our contribution in this work can be summarized as
\begin{itemize}
    \item We introduce a novel approach that integrates constraint learning directly within the imitation learning framework. Unlike existing methods that typically rely on simulators and operate within reinforcement learning or inverse reinforcement learning frameworks, our approach learns constraints from demonstration data. This avoids the complexities of simulator design and eliminates the reality gap between simulated and real-world environments. 
    \item Our constraint module learns driving constraints directly from vectorized scene embeddings. These embeddings efficiently encode critical spatial and temporal relationships in the environment, such as road boundaries, vehicle positions, and obstacle locations, in a structured format. This representation enables the model to effectively identify and generalize constraints across diverse driving scenarios. We formulate the constraint learning problem as a maximum entropy model, which assigns scores to trajectories generated by the motion planner. By decoupling this scoring mechanism into two distinct sub-modules—reward and constraint—we enhance the interpretability of the motion planner’s behavior. Our approach not only improves the model’s visual attention but also reduces risk factors, leading to more transparent and informed decision-making processes. 
    \item We assess the effectiveness of the proposed method using the InD \cite{bock2020ind} and TrafficJams datasets. The experimental results show that integrating constraint learning into the IL framework improves the closed-loop performance of the model.
\end{itemize}

\section{Problem Description and Methodology}

In this section, we present the constraint learning problem. Before delving into its details, we provide a concise overview of our planning pipeline, as we intend to integrate the constraint learning problem into our planning framework.

\subsection{Planning Pipeline}\label{planner-description}

The objective of the planning problem is to find a policy through imitating human driving behaviors. Fig. \ref{fig:architecture} illustrates the architecture of our planner, which is adapted from  \cite{TnT}, originally designed for trajectory prediction. To tailor this framework for planning purposes, we incorporate several modifications and enhancements to suit closed-loop decision-making in autonomous driving. Our model utilizes VectorNet \cite{vectornet} to encode various input features, which include dynamic traffic objects and static map elements. For dynamic features, we consider the ego vehicle and surrounding traffic agents, capturing their spatial and kinematic attributes such as position, velocity, orientation, and size. The map-based static features include centerlines, lane types, speed limits, and other high-definition map data, which are critical for contextual awareness in driving scenarios. The VectorNet architecture is composed of two core components: polyline subgraphs and a global interaction graph. The polyline subgraphs are designed to encode local features from the road and agent elements, such as individual lanes or vehicle trajectories, by representing them as a sequence of connected line segments. This allows the network to capture fine-grained structural information from the environment. On the other hand, the global interaction graph aggregates the information from all polyline subgraphs to model interactions between agents and their surrounding road structures. This is essential for understanding the broader driving context, such as interactions at intersections or lane changes influenced by neighboring vehicles. Once the input features are encoded, the output of the global graph is passed to a target selection module. This module predicts a set of potential future positions for the ego vehicle within a predefined time horizon, using a mixture of learned behaviors and map-based constraints. The targets represent feasible goals for the vehicle, such as specific points on the road or future locations within traffic. Subsequently, these predicted target positions are processed by a motion estimation module. Here, we generate a diverse set of candidate trajectories for the ego vehicle, each representing a potential path from its current position to one of the predicted target locations. To ensure that the trajectories are both feasible and physically plausible, we employ a kinematic bicycle model, which simulates the dynamics of a vehicle during motion planning. This model takes into account the physical constraints of the vehicle, such as turning radius and acceleration limits, and applies a smoothing filter to adjust the trajectory parameters. As a result, the final set of trajectories adheres to both the vehicle's motion constraints and environmental requirements.

\begin{figure}[t]
    \centering
    \includegraphics[trim={0.4cm 4cm 0.4cm 0.4cm}, width=\linewidth]{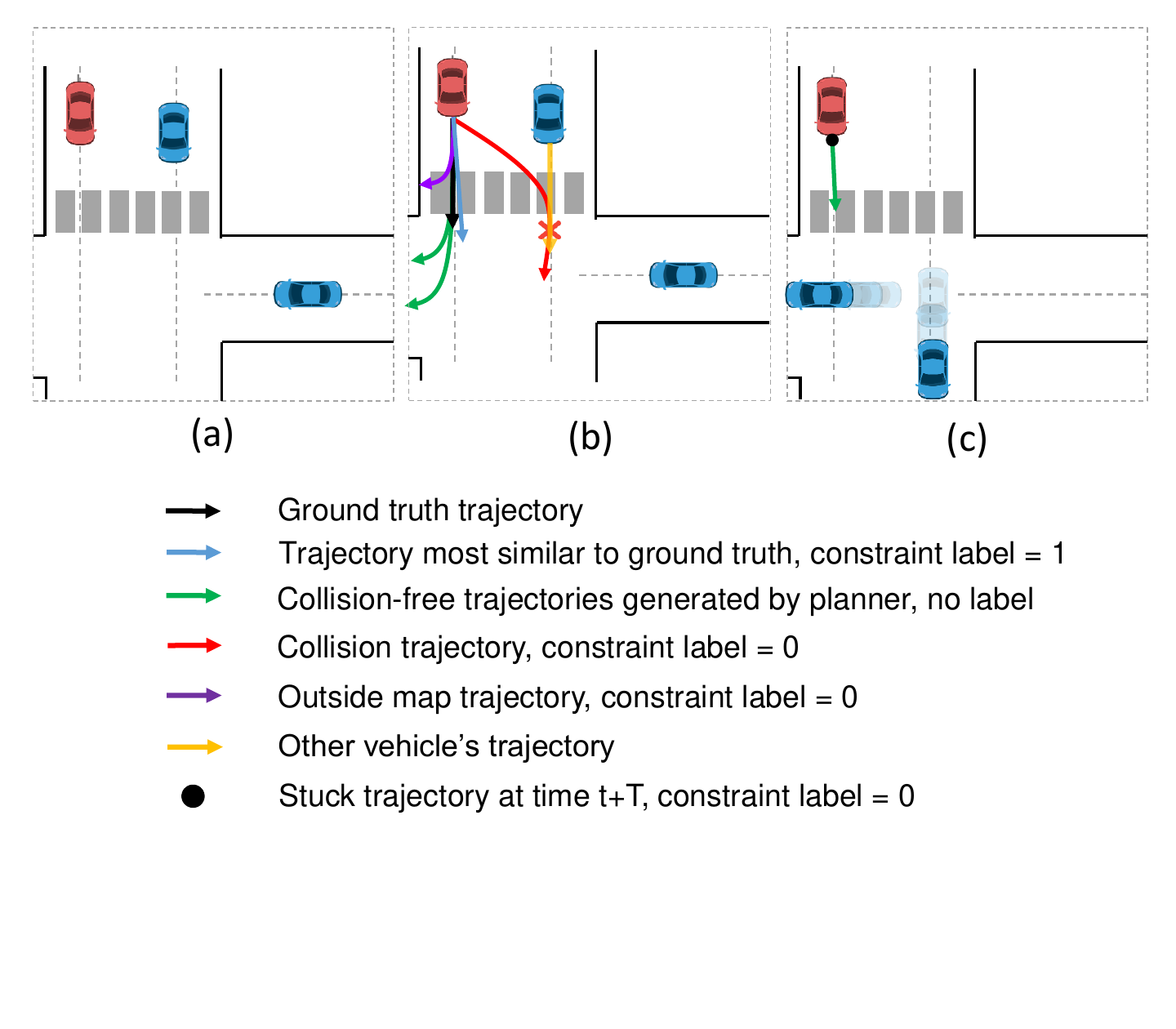}
    \caption{Our constraint labeling scheme: (b) collision and out-of-map trajectories, (c) stuck trajectories. Figure (a) shows a snapshot of the scene at time $t$. In (b), we present different kind of trajectories available to the ego vehicle at time $t$. Figure (c) shows the same scene at time $t+T$. It is worth noting that we label a stuck trajectory only if there exists at least another non-stuck and safe trajectory.}
    \label{fig:labeling}
\end{figure}

\subsection{Integrating Constraints into the Planner Architecture}
As discussed earlier, the objective of our planner is to generate trajectories that closely resemble those of an expert driver. To achieve this, we need a scoring mechanism that assigns higher score values to trajectories exhibiting higher proximity to the expert trajectory. The existing planner architectures take into account score values exclusively based of their resemblance to the expert's trajectory (Fig. \ref{fig:architecture}(a)), without offering insights into the interpretable driving behavior associated with these trajectories. To tackle this limitation, we divide our scoring into two components: reward and constraint, as depicted in Fig. \ref{fig:architecture}(b).

Let $\mathbf{T}$ represent the set of all possible trajectories generated by the motion estimation module. We express the reward function as $r:\mathbf{T}\rightarrow \mathbb{R}$. Also, we denote the constraint function as $c:\mathbf{T}\rightarrow [0,1]$. Using this notation, a constraint value of $0$ indicates that the corresponding trajectory fails to exhibit the desired behavior, violating established driving metrics. For instance, if we define the constraint with respect to collisions, $c(\tau)=0$ indicates that the trajectory $\tau$ results in a collision. Conversely, a constraint value of $1$ asserts that the corresponding trajectory adheres to driving requirements, constituting a safe trajectory. 
As mentioned in section \ref{planner-description}, our motion estimation generates a set of trajectories at each time. Let $N$ denote the number of these trajectories. Also, let $\pi(\cdot)$ show the deployed policy by the planner. To score these trajectories, the original implementation in \cite{TnT} uses a maximum entropy model, where the probability of selecting trajectory $\tau$ is given by
\begin{equation}
\label{score-probability}
P(\tau|\pi) = \frac{e^{r(\tau)}}{\sum_{i=1}^{N} e^{r(\tau_i)}}.
\end{equation}
However, this scoring scheme does not provide any insight into the driving behavior of associated trajectories. To improve interpretability of the model, we modify \eqref{score-probability} such that it considers measure of safety in addition to resemblance to the expert trajectory when scoring the trajectories. Given that $c(\tau)\in[0,1], \forall \tau\in\mathbf{T}$, a natural approach to integrate the constraints into the planning problem is by weighting the selection probabilities of trajectories based on their constraint values. When a trajectory exhibits a low constraint value, it implies undesired performance, requiring a reduction in its selection probability. Conversely, trajectories exhibiting higher quality should correspond to elevated constraint values, necessitating an adjustment in the probability formulation as outlined in \eqref{score-probability}. As a result, following \cite{icml}, we can write the probabilities as
\begin{equation}
\label{score-probability-revised}
P(\tau|\pi) = \frac{ c(\tau) e^{r(\tau)}}{\sum_{i=1}^{N} c(\tau_i) e^{r(\tau_i)}}.
\end{equation}



 



To train the constraint model effectively, it is imperative to establish a suitable loss function. This begins with defining a metric for similarity between two trajectories. Let $s(\tau_1, \tau_2)$ represent the similarity function between two trajectories $\tau_1$ and $\tau_2$, defined as
\begin{equation*}
s(\tau_1, \tau_2) = \lVert \tau_1 - \tau_2\rVert^2.
\end{equation*}
Utilizing this metric, we identify the trajectory closest to the ground truth trajectory as
\begin{equation}
\label{best-traj}
    \tau_{best} = \argmin_{\tau \in \mathcal{T}} s(\tau, \tau_{gt}),
\end{equation}
where $\mathcal{T}$ denotes the set of $N$ trajectories generated by motion estimation, and $\tau_{gt}$ represents the ground truth trajectory. Additionally, let $\bar{\mathcal{T}}$ denote the set of trajectories in $\mathcal{T}$ that contravene driving rules.
Our objective is to learn constraint values such that trajectories similar to the ground truth tend towards a constraint value of $1$, while the constraint values of violating trajectories in $\bar{\mathcal{T}}$ converge to $0$. To achieve this, trajectories violating driving rules are labeled as $0$. The trajectory most similar to the ground truth is assigned a label of $1$. Furthermore, all other trajectories meeting driving rules are not labeled. Fig. \ref{fig:labeling} illustrates the labeling scheme in three different cases, where collision, out-of-map, and stuck serve as the constraint metrics. For the collision and out-of-map, we mark trajectories leading to a future collision or out-of-map behaviour with label of $0$. For the stuck scenario, it is a bit tricky. We record the amount of time the vehicle stays stationary. Once this time exceeds $T$, we assess the available trajectories at time $t+T$. If there exists at least one safe and non-stationary trajectory, we label all stuck trajectories with label of $0$. This is to ensure that cases like staying in traffic jams or behind the red light traffic are not labelled as invalid trajectories. Using this scheme, we define the constraint loss as 
\begin{equation}
\label{constraint-loss}
\mathcal{L}(\theta) = \frac{1}{|\bar{\mathcal{T}}|+1}\sum_{\tau_i \in \bar{\mathcal{T}}} \gamma(c_\theta(\tau_i), 0) + \gamma(c_\theta(\tau_{best}), 1) ,
\end{equation}
where $\theta$ is the parameter of the constraint network and $\gamma(x,y)$ is the cross-entropy loss.
Training the constraint model using $\mathcal{L}(\theta)$ enables the learning of the implicit constraint function. Algorithm I outlines the detailed steps for learning the constraint function.

\begin{algorithm}[h]
\label{constraint-learning-alg}
\caption{Constraint learning algorithm}
\begin{algorithmic}
\State Initialize the constraint model, $c_{\theta}(.)$
\While {not converged}
\State Generate $N$ trajectories using motion estimator
\State Find $\tau_{best}$ using \eqref{best-traj}
\State Set target label of $\tau_{best}$ as $1$
\State Set labels of violating trajectories in $\bar{\mathcal{T}}$ as $0$ 
\State Evaluate the gradient of loss $\mathcal{L}(\theta)$ in \eqref{constraint-loss}
\State Take one step of gradient decent update using 
\[\theta\leftarrow\theta-\eta\nabla_\theta \mathcal{L}(\theta)\]
\EndWhile
\end{algorithmic}
\end{algorithm}



\section{Evaluation and Results}


\subsection{Implementation Setup and Dataset}

\noindent\textbf{Dataset:} We trained and evaluated our proposed constraint learning method using the InD dataset \cite{bock2020ind} and the TrafficJams dataset. The InD dataset contains several hours of measurement data collected at various unsignalized intersections, including trajectories of over 11,500 road users, such as vehicles, cyclists, and pedestrians. The TrafficJams dataset focuses on highway driving scenarios, specifically emphasizing congestion and stop-and-go traffic conditions. 

\begin{figure}[t]
    \centering
    \begin{subfigure}{.49\columnwidth}
        \centering
        \includegraphics[width=\linewidth]{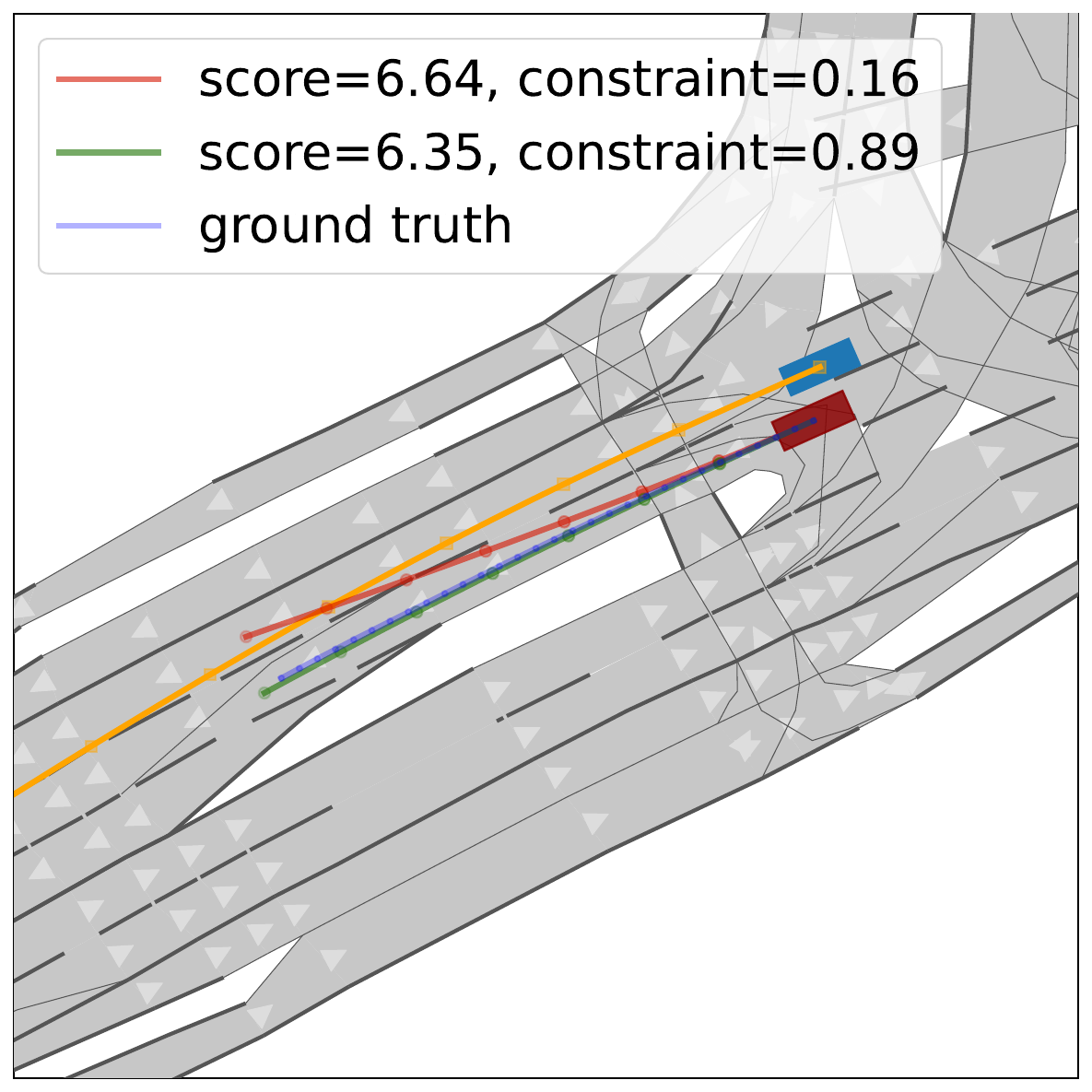}
        \caption{}
        \label{fig_score_same:sub1}
    \end{subfigure}
    \hfill
    \begin{subfigure}{.49\columnwidth}
        \centering
        \includegraphics[width=\linewidth]{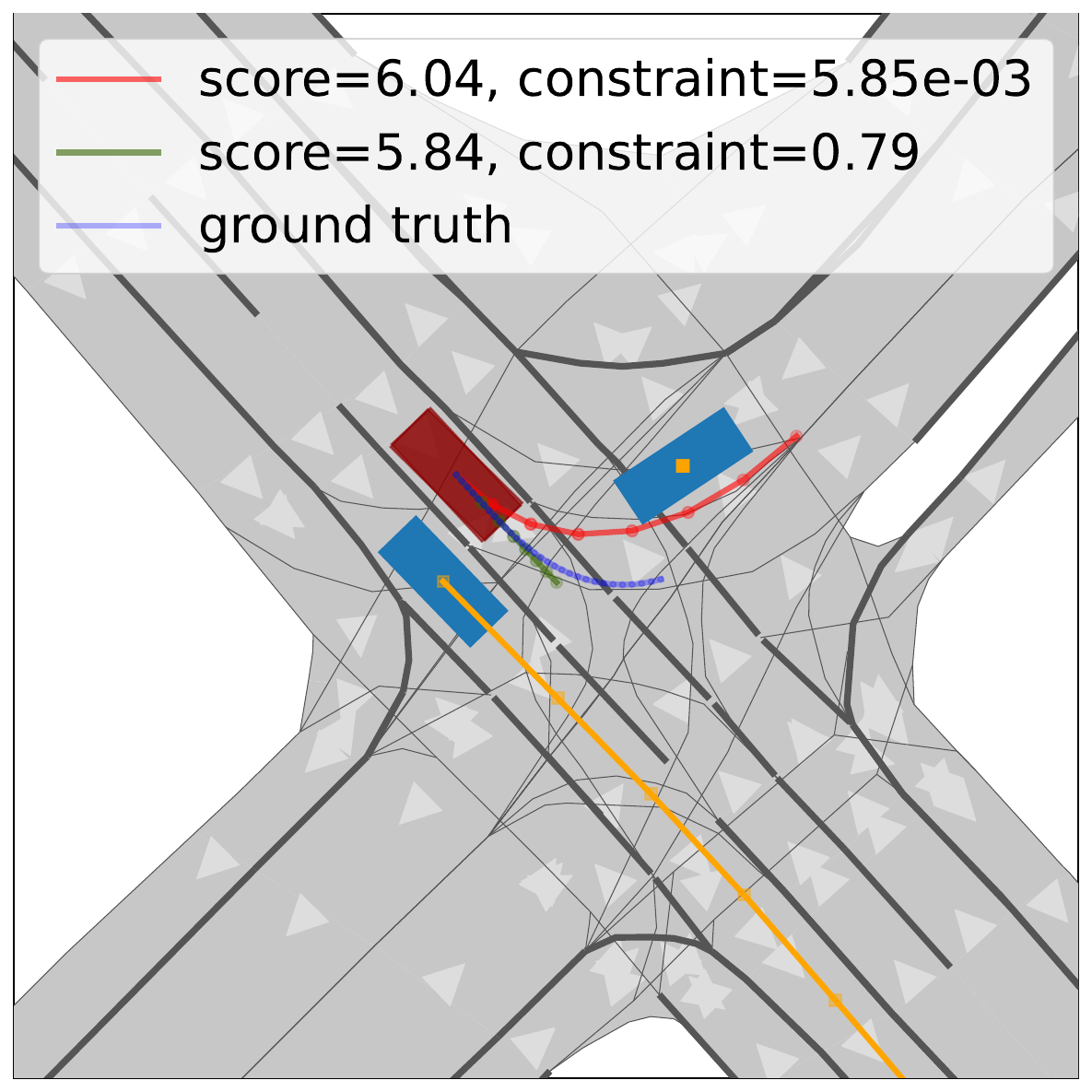}
        \caption{}
        \label{fig_score_same:sub2}
    \end{subfigure}
    \caption{ Snapshots of the scene demonstrating instances where trajectories with higher scores exhibit unsafe behavior, while constraint values effectively distinguish between trajectories. (a) Forward navigation scenario, (b) Left turn at intersection. In both cases, the ego vehicle is shown in red and the traffic vehicles are blue.}
    \label{fig_score_same}
\end{figure}

\begin{table*}[t]
\centering
\caption{Closed-loop performance of the models trained with different constraints, averaged over 5 different seeds.}
\label{table_trained_constraint}
\begin{tabularx}{0.88 \textwidth}{>{\raggedright\arraybackslash}m{2cm}|>{\raggedright\arraybackslash}m{6cm}|>{\centering\arraybackslash}m{1cm}>{\centering\arraybackslash}m{1cm}>{\centering\arraybackslash}m{1cm}>{\centering\arraybackslash}m{1cm}>{\centering\arraybackslash}m{1cm}| X}
\hline
\noalign{\hrule height 0.05cm}
\textbf{Dataset} & \textbf{Experiment} & \textbf{Progress rate (\%)} & \textbf{Success rate (\%)} & \textbf{Collision rate (\%)}   & \textbf{MDBC (m)} & \textbf{outside road (\%)} & \textbf{Risk Factor} \\
\hline
\noalign{\hrule height 0.01cm}
\multirow{4}{2cm}{\textbf{InD}} 
& baseline - no constraints & 84.7 & 60.3 & 29.6  & 194.0 & 0.4 & 0.175  \\
& collision constraint & 85.7 &  67.0 & 22.9 &  
255.9 & 0.2 & 0.159 \\
& collision + out-of-map constraints  & 86.1 & 67.4 & 23.1 & 250.0 & 0.0  & 0.161 \\
& collision + out-of-map + stuck constraints & 87.9 & 69.2 & 24.6  & 252.3 & 0.0 & 0.165 \\
\hline
\noalign{\hrule height 0.01cm}
\multirow{4}{2cm}{\textbf{TrafficJams}} &  baseline - no constraints& 84.79 & 61.69 & 36.94 & 859.88 & 0.3 & 0.214 \\ 
& collision constraint  & 86.68 & 68.64 & 30.50 & 1051.89 & 0.1 & 0.189 \\ 
& collision + out-of-map constraints  & 86.52 & 66.92 & 32.62 & 1043.4 & 0.0 & 0.178 \\
& collision + out-of-map + stuck constraints & 86.82 & 67.79 & 27.11 & 1187.19 & 0.0 & 0.172 \\ 
\hline
\noalign{\hrule height 0.05cm}
\end{tabularx}
\label{tab:grouped_results}
\end{table*}

\noindent\textbf{Training:} For the VectorNet and target selection modules, we used the same implementation as introduced in \cite{TnT}. For the motion estimation, reward, and constraint networks, we use multi-layer perceptrons (MLPs) with two hidden layers of size $64$, followed by the ReLU activation function. Layer normalization is also applied after each hidden layer. The motion estimation calculates the necessary acceleration and rate of turn for the ego vehicle to reach its intended targets. The resulting output from motion estimation is then fed into a bicycle filter to produce the corresponding trajectories. The reward and constraint modules take these trajectories along with the scene embedding to calculate associated reward and constraint values. In order to keep the constraints in the range of $0$ and $1$, the output of constraint network is fed into a Sigmoid activation function. For the constraint labels, we consider all three metrics, including collision, out of map, and stuck. We trained our model for $70$ epochs on Nvidia RTX 2080 Ti GPU. The batch size is $128$ and each training takes approximately $16$ hours.

\noindent\textbf{Evaluation Metrics:}
The evaluation metrics we employ include progress rate, success rate, collision rate, outside road, MDBC, and risk factor. The progress rate expresses the ratio of the distance driven by the vehicle towards the goal compared to the ground truth. Success rate denotes the proportion of scenarios where the ego vehicle successfully reaches its destination without any collisions. Collision rate represents the ratio of scenarios resulting in a collision. Outside road represents the percent of trajectories that invade the road boundaries, and MDBC represents the mean drive between collisions. We also use the risk factor model introduced in \cite{icra24-liu2024safety}, which maps a scenario to a risk score based on the number of unsafe decisions available to the vehicle by the planner. Specifically, the risk factor is calculated as the mean ratio of collision trajectories to total trajectories for each scenario in closed-loop evaluation. This metric offers deeper insights into the safety performance of the system under various conditions. 


\subsection{Results and Discussion}



One advantage of using constraints over traditional scoring methods, as outlined in \cite{TnT}, is that the constraint value provides a direct and interpretable measure of the safety associated with a given trajectory. This is particularly important in safety-critical systems, where understanding and controlling the risk of a trajectory is crucial. In contrast, score values, which are often used to rank trajectories based on various performance metrics, fail to exhibit consistent or reliable patterns across different scenarios, especially when safety is considered. Fig. \ref{fig_score_same} illustrates instances where trajectories with higher scores lead to unsafe outcomes such as collisions. The inconsistency arises because the scoring does not inherently account for safety. As a result, score values alone are unable to distinguish between safe and unsafe trajectories reliably. On the other hand, constraint values effectively distinguish between the trajectories that violate safety conditions and those that do not. Consequently, since the planner typically selects trajectories with the highest scores for future planning, it often lacks the necessary mechanisms to evaluate and ensure the safety of the selected trajectories. By incorporating a constraint module into the planner, the system gains the ability to assess the safety implications of its chosen actions more accurately.

\begin{figure*}[t]
    \centering
    \includegraphics[trim={0cm 0 0cm 0}, clip,  width=\linewidth]{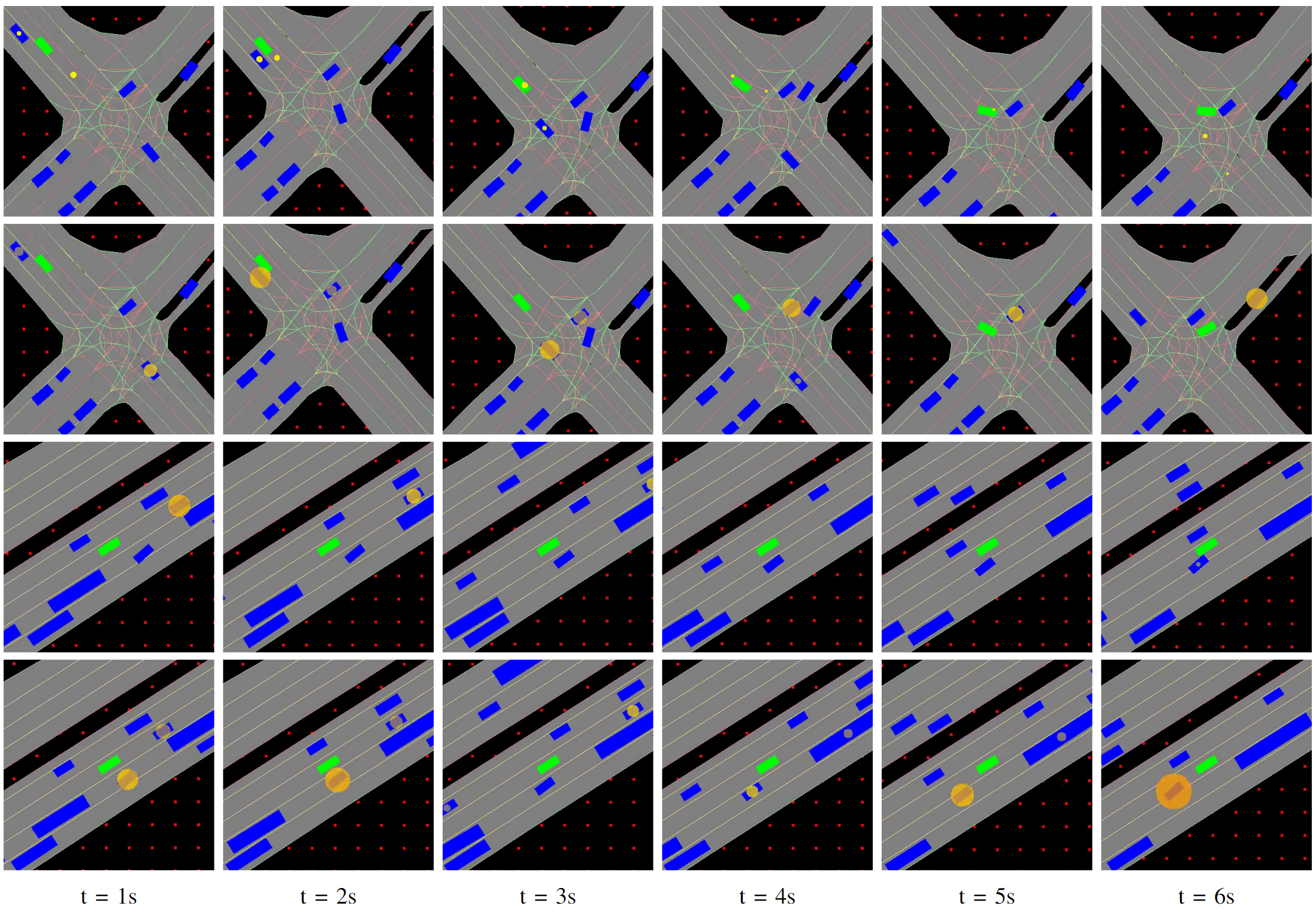}
    \caption{Visualization of the model's attention to objects in the scene. For clarity, only the two objects with the highest attention weights are highlighted, with larger circles indicating greater attention. The top two rows correspond to a scenario from the InD dataset, while the bottom two rows represent a scenario from the TrafficJams dataset. In each scenario, the top row (rows 1 and 3) shows the baseline model, and the bottom row (rows 2 and 4) displays the constraint-based model. As illustrated, the baseline model fails to assign sufficient attention to the causal agents. In contrast, the model trained with constraint modules successfully identifies the causal agents and assigns greater attention to them. Additional videos demonstrating these scenarios and the model's performance are available online ( \href{https://youtu.be/PY0luaE3wYI}{https://youtu.be/PY0luaE3wYI}.)}
\label{fig:attention}
\end{figure*}

Table \ref{table_trained_constraint} presents the closed-loop performance of our model after integrating various constraint modules for both InD and TrafficJams datasets. Initially, we introduced a collision avoidance constraint, and subsequently, we incorporated two additional constraints: the out-of-map constraint and the stuck constraint. The out-of-map constraint was designed to prevent the vehicle from crossing lane boundaries, ensuring that it stays within the intended driving area. This constraint acts as a safeguard, ensuring that the vehicle does not drift off the road or enter restricted zones. Similarly, the stuck constraint was introduced to address situations where the vehicle becomes stationary and unable to proceed, which could cause it to halt prematurely, thereby failing to complete the scenario. The experimental results highlight several positive outcomes from integrating these constraints. For example, the inclusion of the out-of-map constraint completely eliminated the issue of vehicles veering off the road, reducing such incidents to zero. Additionally, the overall success rate of the planner improved, as did its ability to avoid collisions. This collision avoidance improvement is also reflected in the risk factor values, as they count for the number of unsafe trajectories generated by the planner. This indicates a significant enhancement in the planner’s capability to select safer and more reliable trajectories. Moreover, the MDBC metric showed notable improvement, reflecting the model's stronger adherence to safe driving behaviors, even in challenging or unfamiliar scenarios. This improvement underscores the model’s ability to navigate complex situations while maintaining safety.

It is important to note that we have only compared our constraint learning model with the baseline. Existing approaches \cite{liu2022benchmarking, lindner2023learning, howtonotdrive, basich2023learning, hu2024long, pascal_icl, scobee2019maximum} typically require interaction with the environment to learn constraint values, which is a key limitation in our context. Unlike these methods, our approach does not assume access to a precise autonomous driving simulator that faithfully replicates the characteristics of the collected dataset. Since such simulators are unavailable, a direct comparison with these methods would be neither feasible nor meaningful in evaluating our model's effectiveness.

To understand why the model trained with constraint learning modules performs better, we observed that these modules significantly enhance the model's ability to focus on key elements in the scene. Specifically, the model can assign more of its attention to the causal agents—those with the greatest impact on the outcome of the scenario. Fig. \ref{fig:attention}  provides sample snapshots from both the InD and TrafficJams scenarios. In these examples, as shown in row 1 and row 3, the base model struggles to allocate sufficient attention to the causal agents. Particularly, it gives its attention to road elements or non-causal components of the scene, resulting in collisions. In contrast, the model trained with constraint modules (row 2 and row 4) successfully identifies these agents and assigns them the appropriate level of attention, allowing it to navigate the scene without issues. This improvement demonstrates the effectiveness of constraint learning in improving the model’s attention. Additional examples are available in the supplementary video submission.

The results suggest that by separating the scoring module into reward and constraint components, the planner can better handle unseen scenes and generalize more effectively. This ultimately demonstrates the value of integrating constraint learning, not only for ensuring safety but also for improving the model's generalization capabilities.


\section{CONCLUSIONS}


In this work, we have integrated constraint learning directly into the imitation learning framework. Our approach utilizes vectorized scene embeddings that capture encode critical spatial and temporal relationships between scene components, such as vehicle positions, road features, obstacle locations, etc. Using these embeddings, our model learns to identify and generalize constraints across different driving scenarios. Our approach is improves the interpretability of the planner’s behavior, resulting in enhanced safety for autonomous vehicles. A key advantage of our method is that it does not depend on complex simulators or reward-based reinforcement learning techniques, which are often required by state-of-the-art methods. We validated our algorithm using two public datasets, InD and TrafficJams. The results demonstrate that decoupling the scoring module into two distinct components—reward and constraint—significantly enhances the model's ability to detect causal agents in the scene, leading to safer trajectory planning. Our findings also suggest that constraint values offer a more reliable measure of safety compared to traditional score-based methods.

\addtolength{\textheight}{-10cm}   




\bibliographystyle{IEEEtran}
\bibliography{references}

\begin{thebibliography}{10}
\providecommand{\url}[1]{#1}
\csname url@rmstyle\endcsname
\providecommand{\newblock}{\relax}
\providecommand{\bibinfo}[2]{#2}
\providecommand\BIBentrySTDinterwordspacing{\spaceskip=0pt\relax}
\providecommand\BIBentryALTinterwordstretchfactor{4}
\providecommand\BIBentryALTinterwordspacing{\spaceskip=\fontdimen2\font plus
\BIBentryALTinterwordstretchfactor\fontdimen3\font minus \fontdimen4\font\relax}
\providecommand\BIBforeignlanguage[2]{{%
\expandafter\ifx\csname l@#1\endcsname\relax
\typeout{** WARNING: IEEEtran.bst: No hyphenation pattern has been}%
\typeout{** loaded for the language `#1'. Using the pattern for}%
\typeout{** the default language instead.}%
\else
\language=\csname l@#1\endcsname
\fi
#2}}

\bibitem{paden2016survey}
B.~Paden, M.~{\v{C}}{\'a}p, S.~Z. Yong, D.~Yershov, and E.~Frazzoli, ``{A Survey of Motion Planning and Control Techniques for Self-driving Urban Vehicles},'' \emph{IEEE Transactions on intelligent vehicles}, vol.~1, no.~1, pp. 33--55, 2016.

\bibitem{chai2019multipath}
Y.~Chai, B.~Sapp, M.~Bansal, and D.~Anguelov, ``{Multipath: Multiple Probabilistic Anchor Trajectory Hypotheses for Behavior Prediction},'' \emph{arXiv preprint arXiv:1910.05449}, 2019.

\bibitem{chen2023end}
L.~Chen, P.~Wu, K.~Chitta, B.~Jaeger, A.~Geiger, and H.~Li, ``{End-to-end Autonomous Driving: Challenges and Frontiers},'' \emph{arXiv preprint arXiv:2306.16927}, 2023.

\bibitem{chen2019deep}
J.~Chen, B.~Yuan, and M.~Tomizuka, ``{Deep Imitation Learning for Autonomous Driving in Generic Urban Scenarios with Enhanced Safety},'' in \emph{2019 IEEE/RSJ International Conference on Intelligent Robots and Systems (IROS)}.\hskip 1em plus 0.5em minus 0.4em\relax IEEE, 2019, pp. 2884--2890.

\bibitem{le2022survey}
L.~Le~Mero, D.~Yi, M.~Dianati, and A.~Mouzakitis, ``{A Survey on Imitation Learning Techniques for End-to-end Autonomous Vehicles},'' \emph{IEEE Transactions on Intelligent Transportation Systems}, vol.~23, no.~9, pp. 14\,128--14\,147, 2022.

\bibitem{lu2023imitation}
Y.~Lu, J.~Fu, G.~Tucker, X.~Pan, E.~Bronstein, R.~Roelofs, B.~Sapp, B.~White, A.~Faust, S.~Whiteson, \emph{et~al.}, ``Imitation is not enough: Robustifying imitation with reinforcement learning for challenging driving scenarios,'' in \emph{2023 IEEE/RSJ International Conference on Intelligent Robots and Systems (IROS)}.\hskip 1em plus 0.5em minus 0.4em\relax IEEE, 2023, pp. 7553--7560.

\bibitem{liu2022benchmarking}
G.~Liu, Y.~Luo, A.~Gaurav, K.~Rezaee, and P.~Poupart, ``{Benchmarking Constraint Inference in Inverse Reinforcement Learning},'' \emph{2023 International Conference on Learning Representations (ICLR) arXiv:2206.09670}, 2022.

\bibitem{howtonotdrive}
K.~Rezaee and P.~Yadmellat, ``{How To Not Drive: Learning Driving Constraints from Demonstration},'' in \emph{2022 IEEE Intelligent Vehicles Symposium (IV)}.\hskip 1em plus 0.5em minus 0.4em\relax IEEE, 2022, pp. 1297--1302.

\bibitem{chou2020learning}
G.~Chou, D.~Berenson, and N.~Ozay, ``{Learning Constraints from Demonstrations},'' in \emph{Algorithmic Foundations of Robotics XIII: Proceedings of the 13th Workshop on the Algorithmic Foundations of Robotics 13}.\hskip 1em plus 0.5em minus 0.4em\relax Springer, 2020, pp. 228--245.

\bibitem{scobee2019maximum}
D.~R. Scobee and S.~S. Sastry, ``{Maximum Likelihood Constraint Inference for Inverse Reinforcement Learning},'' in \emph{International Conference on Learning Representations (ICLR)}, 2019.

\bibitem{icml}
S.~Malik, U.~Anwar, A.~Aghasi, and A.~Ahmed, ``{Inverse Constrained Reinforcement Learning},'' in \emph{Proceedings of the 38th International Conference on Machine Learning (ICML)}, 2021, pp. 7390--7399.

\bibitem{pascal_icl}
A.~Gaurav, K.~Rezaee, G.~Liu, and P.~Poupart, ``{Learning Soft Constraints from Constrained Expert Demonstrations},'' \emph{International Conference on Learning Representations (ICLR) arXiv:2206.01311}, 2022.

\bibitem{kim2023learning}
K.~Kim, G.~Swamy, Z.~Liu, D.~Zhao, S.~Choudhury, and Z.~S. Wu, ``{Learning Shared Safety Constraints from Multi-task Demonstrations},'' \emph{arXiv preprint arXiv:2309.00711}, 2023.

\bibitem{wang2024large}
J.~Wang, Z.~Wu, Y.~Li, H.~Jiang, P.~Shu, E.~Shi, H.~Hu, C.~Ma, Y.~Liu, X.~Wang, \emph{et~al.}, ``Large language models for robotics: Opportunities, challenges, and perspectives,'' \emph{arXiv preprint arXiv:2401.04334}, 2024.

\bibitem{hu2024long}
X.~Hu, P.~Chen, Y.~Wen, B.~Tang, and L.~Chen, ``Long and short-term constraints driven safe reinforcement learning for autonomous driving,'' \emph{arXiv preprint arXiv:2403.18209}, 2024.

\bibitem{wang2024safe}
C.~Wang and Y.~Wang, ``Safe autonomous driving with latent dynamics and state-wise constraints,'' \emph{Sensors}, vol.~24, no.~10, p. 3139, 2024.

\bibitem{ziebart2008maximum}
B.~D. Ziebart, A.~L. Maas, J.~A. Bagnell, A.~K. Dey, \emph{et~al.}, ``{Maximum Entropy Inverse Reinforcement Learning},'' in \emph{23rd AAAI Conference on Artificial Intelligence (AAAI)}, vol.~8.\hskip 1em plus 0.5em minus 0.4em\relax Chicago, IL, USA, 2008, pp. 1433--1438.

\bibitem{vitelli2022safetynet}
M.~Vitelli, Y.~Chang, Y.~Ye, A.~Ferreira, M.~Wo{\l}czyk, B.~Osi{\'n}ski, M.~Niendorf, H.~Grimmett, Q.~Huang, A.~Jain, \emph{et~al.}, ``Safetynet: Safe planning for real-world self-driving vehicles using machine-learned policies,'' in \emph{2022 International Conference on Robotics and Automation (ICRA)}.\hskip 1em plus 0.5em minus 0.4em\relax IEEE, 2022, pp. 897--904.

\bibitem{lindner2023learning}
D.~Lindner, X.~Chen, S.~Tschiatschek, K.~Hofmann, and A.~Krause, ``{Learning Safety Constraints from Demonstrations with Unknown Rewards},'' \emph{arXiv preprint arXiv:2305.16147}, 2023.

\bibitem{basich2023learning}
C.~Basich, S.~Mahmud, and S.~Zilberstein, ``Learning constraints on autonomous behavior from proactive feedback,'' in \emph{2023 IEEE/RSJ International Conference on Intelligent Robots and Systems (IROS)}.\hskip 1em plus 0.5em minus 0.4em\relax IEEE, 2023, pp. 3680--3687.

\bibitem{highwayenv}
E.~Leurent, ``{An Environment for Autonomous Driving Decision-Making},'' \url{https://github.com/eleurent/highway-env}, 2018.

\bibitem{bock2020ind}
J.~Bock, R.~Krajewski, T.~Moers, S.~Runde, L.~Vater, and L.~Eckstein, ``{The Ind Dataset: A Drone Dataset of Naturalistic Road User Trajectories at German Intersections},'' in \emph{2020 IEEE Intelligent Vehicles Symposium (IV)}.\hskip 1em plus 0.5em minus 0.4em\relax IEEE, 2020, pp. 1929--1934.

\bibitem{TnT}
H.~Zhao, J.~Gao, T.~Lan, C.~Sun, B.~Sapp, B.~Varadarajan, Y.~Shen, Y.~Shen, Y.~Chai, C.~Schmid, \emph{et~al.}, ``{Tnt: Target-Driven Trajectory Prediction},'' in \emph{Conference on Robot Learning}.\hskip 1em plus 0.5em minus 0.4em\relax PMLR, 2021, pp. 895--904.

\bibitem{vectornet}
J.~Gao, C.~Sun, H.~Zhao, Y.~Shen, D.~Anguelov, C.~Li, and C.~Schmid, ``{Vectornet: Encoding HD Maps and Agent Dynamics from Vectorized Representation},'' in \emph{Proceedings of the IEEE/CVF Conference on Computer Vision and Pattern Recognition}, 2020, pp. 11\,525--11\,533.

\bibitem{icra24-liu2024safety}
H.~Liu, L.~Zhang, S.~K.~S. Hari, and J.~Zhao, ``{Safety-Critical Scenario Generation Via Reinforcement Learning Based Editing},'' in \emph{2024 IEEE International Conference on Robotics and Automation (ICRA)}.\hskip 1em plus 0.5em minus 0.4em\relax IEEE, 2024, pp. 14\,405--14\,412.

\end{thebibliography}

\end{document}